\documentclass[runningheads]{llncs}
\usepackage{graphicx}

\usepackage{tikz}
\usepackage{comment}
\usepackage{amsmath,amssymb} 
\usepackage{color}
\usepackage{xspace}
\usepackage{orcidlink}
\usepackage{graphicx}
\usepackage{bm, mathrsfs}
\usepackage{booktabs}
\usepackage[nice]{nicefrac}
\usepackage{adjustbox}
\usepackage{multirow}
\usepackage{pifont}
\usepackage{listings}
\usepackage{xcolor}
\usepackage{enumitem}
\usepackage{makecell}
\usepackage{wrapfig}
\usepackage{algorithmic}
\usepackage{algorithm}
\usepackage[font=small,labelfont=bf,tableposition=top]{caption}

\DeclareCaptionLabelFormat{andtable}{#1~#2  \&  \tablename~\thetable}

\hypersetup{colorlinks, breaklinks, citecolor=[rgb]{0., 0., 0.},
anchorcolor=[rgb]{0.,0.,0.}, linkcolor=[rgb]{0.0, 0., 0.},
urlcolor=[rgb]{0., 0., 0.}
}

\usepackage[accsupp]{axessibility}  

\newcommand{\bftab}{\fontseries{b}\selectfont}

\begin{document}
\pagestyle{headings}
\mainmatter
\def\ECCVSubNumber{601}  

\makeatletter
\DeclareRobustCommand\onedot{\futurelet\@let@token\@onedot}
\def\@onedot{\ifx\@let@token.\else.\null\fi\xspace}
\def\eg{\emph{e.g}\onedot} \def\Eg{\emph{E.g}\onedot}
\def\ie{\emph{i.e}\onedot} \def\Ie{\emph{I.e}\onedot}
\def\cf{\emph{cf}\onedot} \def\Cf{\emph{Cf}\onedot}
\def\etc{\emph{etc}\onedot} \def\vs{\emph{vs}\onedot}
\def\wrt{w.r.t\onedot} \def\dof{d.o.f\onedot}
\def\iid{i.i.d\onedot} \def\wolog{w.l.o.g\onedot}
\def\etal{\emph{et al}\onedot}

\lstset{
  backgroundcolor=\color{white},
  basicstyle=\fontsize{7.5pt}{7.5pt}\ttfamily\selectfont,
  columns=fullflexible,
  breaklines=true,
  captionpos=b,
  commentstyle=\fontsize{7.5pt}{7.5pt}\color{codeblue},
  keywordstyle=\fontsize{7.5pt}{7.5pt}\color{codekw},
}

\title{Neuromorphic Data Augmentation for Training Spiking Neural Networks} 


\titlerunning{Neuromorphic Data Augmentation}
%
\author{Yuhang Li\orcidlink{0000-0002-6444-7253} \and
Youngeun Kim\orcidlink{0000-0002-3542-7720} \and
Hyoungseob Park\orcidlink{0000-0003-0787-2082} \and Tamar Geller\orcidlink{0000-0001-8358-1784} \and \\ Priyadarshini Panda\orcidlink{0000-0002-4167-6782}}
\authorrunning{Y. Li, Y. Kim, H. Park, T. Geller, P. Panda.}
%
\institute{Yale University, New Haven, CT 06511, USA\\
\email{\scriptsize\{yuhang.li,youngeun.kim,hyoungseob.park,tamar.geller,priya.panda\}@yale.edu}}
\maketitle

\begin{abstract}
Developing neuromorphic intelligence on event-based datasets with Spiking Neural Networks (SNNs) has recently attracted much research attention. However, the limited size of event-based datasets makes SNNs prone to overfitting and unstable convergence. This issue remains unexplored by previous academic works.
In an effort to minimize this generalization gap, we propose Neuromorphic Data Augmentation (NDA), a family of geometric augmentations specifically designed for event-based datasets with the goal of significantly stabilizing the SNN training and reducing the generalization gap between training and test performance.
The proposed method is simple and compatible with existing SNN training pipelines. Using the proposed augmentation, for the first time, we demonstrate the feasibility of unsupervised contrastive learning for SNNs. We conduct comprehensive experiments on prevailing neuromorphic vision benchmarks and show that NDA yields substantial improvements over previous state-of-the-art results. For example, the NDA-based SNN achieves accuracy gain on CIFAR10-DVS and N-Caltech 101 by 10.1\% and 13.7\%, respectively.
Code is available on GitHub (\href{https://github.com/Intelligent-Computing-Lab-Yale/NDA_SNN}{URL}). 
\keywords{Data Augmentation, Event-based Vision, Spiking Neural Networks}
\end{abstract}

\section{Introduction}
\label{sec:intro}

\begin{figure}[t]
    \centering
    \includegraphics[width=\linewidth]{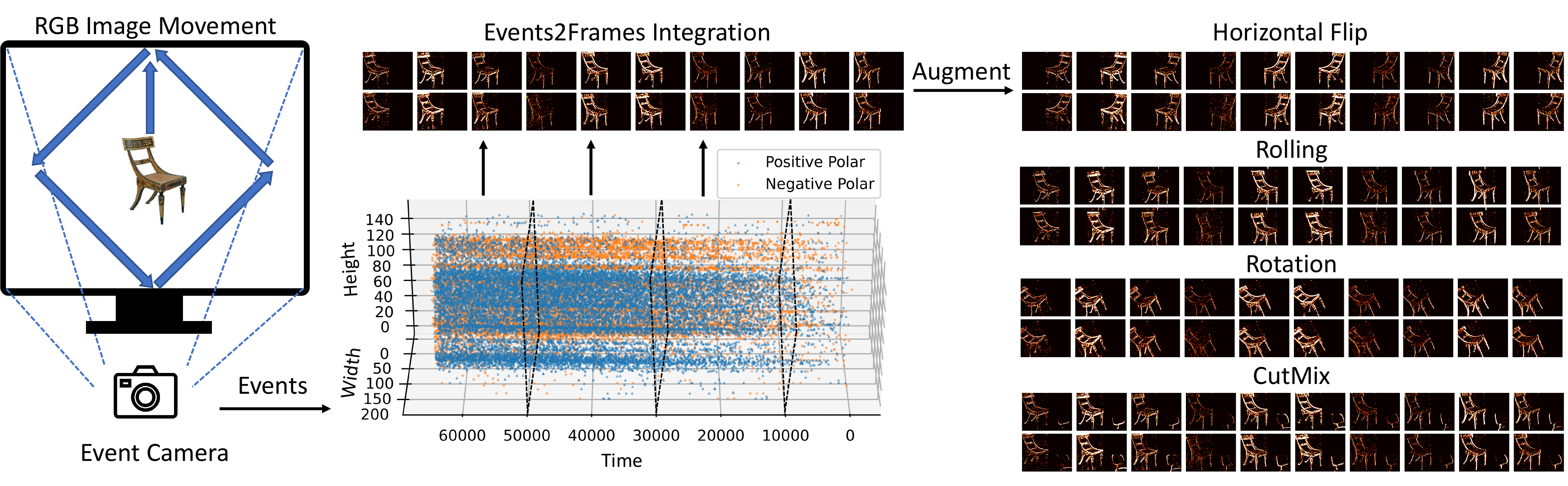}
    \caption{\textit{Left}: Event data collection of a chair image from N-Caltech101~\cite{orchard2015converting}, these events are too sparse to process, thus we integrate them into frames of sparse tensors. \textit{Right}: Our neuromorphic data augmentation on events data.}
    \label{fig_intro}
\end{figure}

Spiking Neural Networks (SNNs), a representative category of models in neuromorphic computing, have received attention as a prospective candidate for low-power machine intelligence \cite{roy2019towards}. Unlike the standard Artificial Neural Networks (ANNs), SNNs deal with binarized spatial-temporal data. A popular example of this data is the DVS event-based dataset.\footnote{In this paper, most event-based datasets we use are collected with Dynamic Vision Sensor (DVS) cameras, therefore we also term them as DVS data for simplicity.} Each pixel inside a DVS camera is operated independently and asynchronously, reporting new brightness when it changes, and staying silent otherwise~\cite{Eventcamera}.
The spatial-temporal information encoded in DVS data can be suitably leveraged by the temporally evolving spiking neurons in SNNs.
In Fig.~\ref{fig_intro}, we describe the process of data recording with a DVS camera.
First, an RGB image is programmed to move in a certain trajectory (or the camera moves in a reverse way), and then the event camera records the brightness change and outputs an event stream. Note that the raw event-based DVS data is too sparse to extract features. Thus, we integrate the events into multiple frames and we study this sparse frame-based data~\cite{wu2018spatio} (see the events2frames integration in Fig.~\ref{fig_intro}).

\begin{figure}[t]
    \centering
    \includegraphics[width=\linewidth]{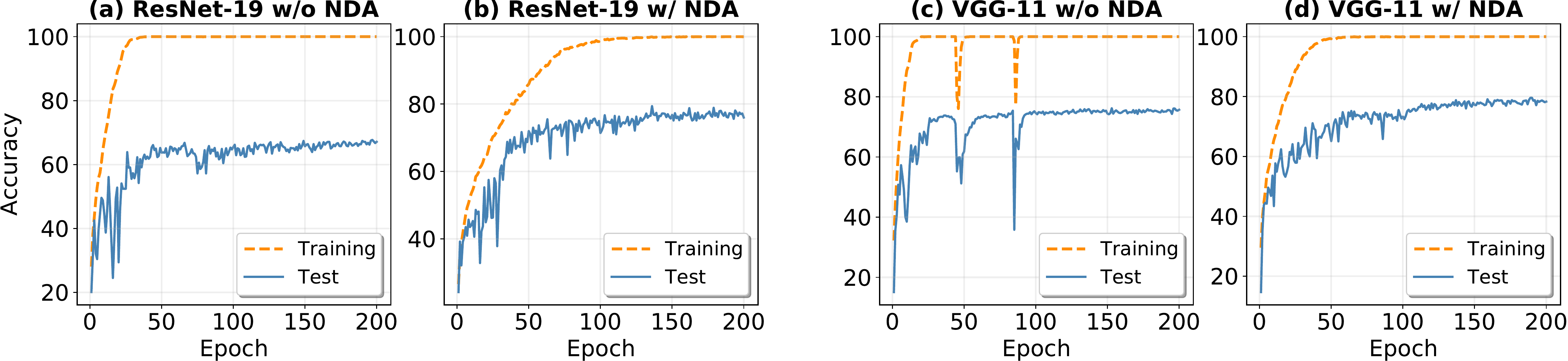}
    \caption{Training/test accuracy curves of ResNet-19 and VGG-11 on CIFAR10-DVS dataset. Networks trained with NDA tends to have better convergence.}
    \label{fig_train_test}
\end{figure}

Due to the high cost of collecting DVS data~\cite{lin2021imagenet}, existing DVS datasets usually contain limited data instances~\cite{li2017cifar10}. 
Consequently, models trained on raw DVS datasets exhibit large training-test performance gaps due to overfitting. 
For instance, CIFAR10-DVS~\cite{li2017cifar10} only contains 10k data points while CIFAR-10 RGB dataset contains 60k data points.
In Fig.~\ref{fig_train_test}, we plot the training-test accuracies on CIFAR10-DVS using VGG-11~\cite{simonyan2014very} and ResNet-19~\cite{he2016deep}.
It is obvious that training accuracy increases swiftly and smoothly. On the contrary, the test accuracy oscillates and remains at half of the training accuracy.
Although advanced training algorithms~\cite{zheng2020going,fang2021incorporating}
have been proposed to improve the generalization of SNNs, data scarcity remains a major challenge and needs to be addressed. 
A naive way to increase DVS data samples is using the DVS camera to record additional augmented RGB images.
However, during training, this carries prohibitively high latency overhead due to the cost of DVS recording. 
Because of this, we focus on a data augmentation technique that can be directly applied to DVS data in order to balance efficiency and effectiveness.

We propose Neuromorphic Data Augmentation (NDA) to transform the off-the-shelf recorded DVS data in a way that prevents overfitting. To ensure the consistency between augmentation and event-stream generation, we extensively investigate various possible augmentations and identify the most beneficial family of augmentations.
In addition, we show that the proposed data augmentation techniques lead to a substantial gain of training stability and generalization performance (cf. Fig.~\ref{fig_train_test}).
Even more remarkably, NDA enables SNNs to be trained through unsupervised contrastive learning without the usage of labels.
Fig.~\ref{fig_intro} (right) demonstrates four examples where NDA is applied. 

The main contributions of this paper are:
\begin{enumerate}[nosep]
    \item[1.] We propose Neuromorphic Data Augmentation for training Spiking Neural Networks on event datasets. Our proposed augmentation policy significantly improves the test accuracy of the model with negligible cost. Furthermore, NDA is compatible with existing training algorithm. 
    \item[2.] We conduct extensive experiments to verify the effectiveness of our proposed NDA on several benchmark DVS datasets like CIFAR10-DVS, N-Caltech 101, N-Cars, and N-MNIST. NDA brings a significant accuracy boost when compared to the previous state-of-the-art results. 
    \item[3.] In a first of its kind analysis, we show the suitability of NDA for unsupervised contrastive learning on the event-based datasets, enabling SNN feature extractors to be trained without labels. 
\end{enumerate}

\section{Related Work}
\subsection{Data Augmentation}

Today, deep learning is a prevalent technique in various commercial, scientific, and academic applications. 
Data augmentation~\cite{shorten2019survey} plays an indispensable role in the deep learning model, as it forces the model to learn invariant features, and thus helps generalization.
The data augmentation is applied in many areas of vision tasks including object recognition~\cite{cubuk2018autoaugment,lim2019fastautoaug,cubuk2020randaugment}, objection detection~\cite{zoph2020learning}, and semantic segmentation~\cite{ronneberger2015u}. 
Apart from learning invariant features, data augmentation also has other specific applications in deep learning. For example, adversarial training~\cite{ganin2016domain,tramer2017ensemble} leverages data augmentation to create adversarial samples and thereby improves the adversarial robustness of the model. Data augmentation is also used in generative adversarial networks (GAN)~\cite{arjovsky2017wasserstein,karras2017progressive,gulrajani2017improved,brock2018large}, neural style transfer~\cite{jing2019neural,gatys2016image}, and data inversion~\cite{zhang2021diversifying,li2021mixmix}.
For event-based data, there are few data augmentation techniques. EventDrop~\cite{gu2021eventdrop}, for example, randomly removes several events due to noise produced in the DVS camera. Our work, on the other hand, tackles the consistency problem of directly applying data augmentations to event-based data. 
A related augmentation technique is video augmentation~\cite{budvytis2017large}, which also augmentation spatial-temporal data. The main difference is that video augmentation can utilize photometric and color augmentations while our NDA can only adopt geometric augmentations.

\subsection{Spiking Neural Networks}

Spiking Neural Networks (SNNs) are often recognized as the third generation of generalization methods for artificial intelligence~\cite{basegmez2014next}. Unlike traditional Artificial Neural Networks (ANNs), SNNs apply the spiking mechanism inside the network. 
Therefore, the activation in SNNs is binary and adds a temporal dimension. Current practices consist of two approaches for obtaining an SNN: \textit{Direct training} and \textit{ANN-to-SNN conversion}. Conversion from a pre-trained ANN can guarantee good performance~\cite{deng2021optimal,diehl2016conversion,rueckauer2016theory,li2021free,li2022converting,han2020deep}, however, in this paper, we do not study conversion since our topic is a neuromorphic dataset that requires direct training. Training SNNs \cite{wu2019direct,wu2018spatio,hazan2018bindsnet,li2021differentiable,deng2022temporal,guo2022recdis} requires spatial-temporal backpropagation. Recently, more methods and variants for such spatio-temporal training have been proposed: \cite{rathi2020enabling} present hybrid training, \cite{zheng2020going,kim2020revisiting} propose a variant of batch normalization~\cite{ioffe2015batch} for SNN;~\cite{rathi2020diet,fang2021incorporating} propose training threshold and potential for better accuracy. However, most of these works focus on algorithm or architecture optimization to achieve improved performance. Because of their sparse nature, DVS datasets present an orthogonal problem of overfitting that is not addressed by the previously mentioned works.

\section{Conventional Augmentations for RGB Data}
\label{sec_da}

We generally divide the current data augmentation techniques for ANN training on natural RGB image datasets into two types, namely photometric \& color augmentations and geometric augmentations. 

\begin{figure}[t]
    \centering
    \includegraphics[width=\linewidth]{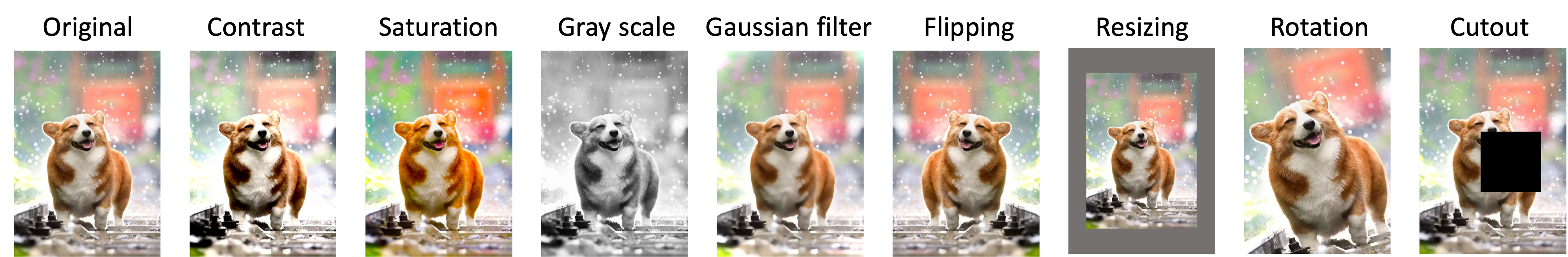}
    \caption{Types of RGB data augmentation. The 2-5th examples include the photometric \& color augmentations and the 6-9th examples contain the geometric augmentations.}
    \label{fig_data_aug}
\end{figure}

\noindent{\bftab Photometric and Color Augmentations. }
This type indicates any augmentations that can transform the image illumination and color space.  Fig.~\ref{fig_data_aug}~(2nd-5th examples) demonstrates some examples of increasing contrast, saturation, and imposing gray scale as well as Gaussian blur. Typically, photometric augmentation is applied to an image by changing each pixel value. For instance, the contrast enhancement will use $f(x) = \mathrm{clip}(ax-\frac{1}{2}a+\frac{1}{2}, 0, 1)$ where $a>1$, to push pixel values close to black (zero value) and white (one value) and is applied in a pixel-wise manner. The color augmentation includes some color space transformation by casting more red, blue, or green colors on the image (\eg saturation and grayscale). 
Generally, both Photometric and Color augmentations can be categorized as value-based augmentation where a transformation $f(x)$ applies to all pixels. Therefore, we also categorize augmentations like a Gaussian filter where a convolution with a Gaussian kernel is applied to this class.  

\noindent{\bftab Geometric Augmentations. }
Unlike \textit{value-based} augmentations,  geometric augmentations do not seek to alter every pixel value with a certain criterion. Rather, they change the images based on the coordinate of each pixel, \ie the index of each element. In Fig.~\ref{fig_data_aug}~(6th-9th examples), we visualize several geometric augmentation examples of horizontal flipping, resizing, rotation and cutout. 
For example, horizontal flipping reverses the order of each pixel and turns the image by 180 degrees. Rotation can be viewed as moving the location of pixels to another place. Cutout~\cite{devries2017cutout} applies a 0/1 mask to the original image. All of these augmentations are \textit{index-based}.

\section{Neuromorphic Augmentations for Event-Based Data}

The event-based datasets contain sparse, discrete, and time-series samples, which have fundamentally different formats when compared to RGB images. 
As a result, the above conventional augmentations cannot all be directly applied. To explore which augmentation is useful for event data, we study the case of photometric and geometric augmentations separately. We also discuss the potential application of neuromorphic data augmentation. 

\subsection{DVS Data and Augmentation}

DVS cameras are data-driven sensors: their output depends on the amount of motion or brightness change in the scene~\cite{gallego2019event}. Mathematically, the output of event camera $\bm{x}_E\in\mathbb{B}^{t\times p\times w\times h}$ is a 4-D tensor ($\mathbb{B}$ is the binary domain $\{0,1\}$ and here we treat it as a sparse binary tensor, \ie we also record 0 if there are no events), where $t$ is the time step, $p$ is the polarity ($p=2$ corresponding to positive and negative polars) and $w,h$ are the width and the height of the event tensor. The event generation (see details in \cite{gallego2019event}) can be modeled as 
{
\begin{equation}
    \bm{x}_E(t, 1, x, y) = \begin{cases}
    1 & \text{if }\mathrm{log}V(t, x, y) - \mathrm{log}V(t-\Delta t,x,y) > \alpha \\
    0 & \text{otherwise}
\end{cases},
\end{equation}
\begin{equation}
    \bm{x}_E(t, 0, x, y) = \begin{cases}
    1 & \text{if }\mathrm{log}V(t, x, y) - \mathrm{log}V(t-\Delta t,x,y) < - \alpha \\
    0 & \text{otherwise}
    \end{cases},
\end{equation}
}where, $\bm{x}_E(t, 1, x, y)$ is the generated event stream at time step $t$, spatial coordinate $(x,y)$ and positive polarity. $V$ is the photocurrent (or ``brightness") of the original RGB image. 
$\Delta t$ is the time elapsed since the last event generated at
the same coordinate.
During DVS data recording, the RGB images are programmed to move in a certain trajectory (or the camera moves in a reverse way). As time evolves, if the pixel value changes fast and exceeds a certain threshold $\alpha$, an event will be generated, otherwise it will stay silent, meaning that the output is 0 in $\bm{x}_E$. The event stream will be split into two channels, \ie two polarities. Positive events are integrated into positive polarity and vice versa.

Consider the RGB data as $\bm{x}\in\mathbb{C}^{3\times w\times h}$ ($\mathbb{C}$ is the continuous domain in $[0, 1]$). We use the function $f(\bm{x}):\mathbb{C}^{3\times w\times h}\rightarrow \mathbb{C}^{3\times w\times h}$ to augment the RGB data. Note that $f(\bm{x})$ can be both photometric or geometric augmentation, and is randomly sampled from a set of augmentations. The optimization objective of training an ANN with RGB augmented data can be given by $\min_{\bm{w}} \frac{1}{n}\sum_{i=1}^n\ell(\bm{w}, f(\bm{x}_i), \bm{y}_i).$

Now consider event-based data $\bm{x}_E$. We define RGB to DVS data function $\bm{x}_E =g(\bm{x}):\mathbb{C}^{3\times w\times h}\rightarrow \mathbb{B}^{t\times p\times w\times h}$.\footnote{Note that using event camera $g(\bm{x})$ to generate DVS data is expensive and impractical during run-time. It is easier to pre-collect the DVS data with a DVS camera and, then work with the DVS data during runtime.}
A naive approach to augment DVS data when training SNN is to first augment RGB data and then translate them to event-stream form, \ie $g\circ f(\bm{x})$.
This method can ensure the augmentation is correctly implemented as well as yield the event-stream data. In fact, training with Poisson encoding~\cite{diehl2015fast,roy2019towards} uses such form $g\circ f(\bm{x})$ where $g$ is the encoding function that translates the RGB images to spike trains.
However, unlike Poisson encoding which can be implemented with one line of code, it would be very time-consuming and expensive to generate a large amount of DVS augmented data, \ie $g\circ f(\bm{x})$. We propose a more efficient method, the neuromorphic data augmentation $f_{\text{NDA}}$ which is directly applied to the DVS data $f_{\text{NDA}}\circ g(\bm{x})$. As a result, we avoid the expensive $g\circ f(\bm{x})$ in the training phase.

\setlength{\tabcolsep}{7pt}
\begin{table}[t]
    \centering
    \caption{Comparison among all potential augmentation for DVS data. $f_{P}$ and $f_G$ mean photometric and geometric augmentations. $\mathbb{C}$ stands for the continuous domain in $[0, 1]$ and $\mathbb{B}$ stands for the binary domain $\{0, 1\}$.}
    \begin{adjustbox}{max width=\linewidth}
    \begin{tabular}{c c  c  l  l}
    \toprule
    {\bftab Aug.} & {\bftab Combination} & {\bftab Input-output} & \multicolumn{1}{c}{{\bftab Pros}} & \multicolumn{1}{c}{{\bftab Cons}} \\
    \noalign{\smallskip}
    \hline
    \noalign{\smallskip}
    $f_{P}$ & $g\circ f_{P}(\bm{x})$ & $\mathbb{C}^{3\times w\times h}\rightarrow \mathbb{B}^{t\times p\times w\times h}$ & \makecell[l]{{\ \ \scriptsize\romannumeral1. Effective} {\scriptsize augmentation}} & \makecell[l]{{ \scriptsize\romannumeral1. Impractical to record} \\ {\scriptsize \ \ \ \ \,huge amount of DVS data}} \\
    \noalign{\smallskip}
    \hline
    \noalign{\smallskip}
    $f_{P}$ & $f_{P}\circ g(\bm{x})$ & $\mathbb{C}^{3\times w\times h}\rightarrow \mathbb{C}^{t\times p\times w\times h}$ & \makecell[l]{\ \ \scriptsize\romannumeral1. Practical} & \makecell[l]{{\ \scriptsize\romannumeral1. Not effective,} \\ {\scriptsize\,\romannumeral2. Creates continuous data}} \\
    \noalign{\smallskip}
    \hline
    \noalign{\smallskip}
    $f_G$ & $g\circ f_G(\bm{x})$ & $\mathbb{C}^{3\times w\times h}\rightarrow \mathbb{B}^{t\times p\times w\times h}$ & \makecell[l]{{\ \ \scriptsize\romannumeral1. Effective} {\scriptsize augmentation}} & \makecell[l]{{ \scriptsize\romannumeral1. Impractical to record} \\ {\scriptsize \ \ \ \ \,huge amount of DVS data}} \\
    \noalign{\smallskip}
    \hline
    \noalign{\smallskip}
    $f_G$ & $f_{G}\circ g(\bm{x})$ & $\mathbb{C}^{3\times w\times h}\rightarrow \mathbb{B}^{t\times p\times w\times h}$ &\makecell[l]{{\ \ \scriptsize\romannumeral1. Practical and effective,} \\{\ \,\scriptsize\romannumeral2. Approximates $g\circ f_G(\bm{x})$}} & {\scriptsize \ \ \ \ \ None} \\
    \bottomrule
    \end{tabular}
    \end{adjustbox}
    \label{tab_augconvert}
\end{table}

Ideally, NDA is supposed to satisfy $f_{\text{NDA}}\circ g(\bm{x})\approx g\circ f(\bm{x})$. To fulfill this commutative law, the NDA data augmentation function must have the mapping of $f_{\text{NDA}}:\mathbb{B}^{t\times p\times w\times h} \rightarrow \mathbb{B}^{t\times p\times w\times h}$.
Without loss of generality, a core component is to evaluate whether an arbitrary augmentation $f(\cdot)$ can achieve 
\begin{equation}
    f(H_\alpha\left[\mathrm{log}V(t) - \mathrm{log}V(t-\Delta t)\right]) \approx H_\alpha\left[ \mathrm{log}f(V(t)) - \mathrm{log}f(V(t-\Delta t))\right],
    \label{eq_condition}
\end{equation}
where $H_\alpha[\cdot]$ is the Heaviside step function (\ie returns 1 when the input is greater than $\alpha$ otherwise 0). Note that we omit the spatial coordinate here.
Recall that photometric \& color augmentation are \textit{value-based} augmentation schemes. This brings two problems: first, on the left hand side of Eq.~(\ref{eq_condition}), the augmented DVS data is not event-stream since the \textit{value-based} transformation outputs continuous values. Second, on the right hand side of Eq.~(\ref{eq_condition}), predicting the event is intractable. Due to brightness difference change, it is unclear whether the event is activated or remains silent after augmentation ($f(\mathrm{log}V(t)) - f(\mathrm{log}V(t-1))$).  
Therefore, we cannot use this type of augmentation in NDA.

It turns out that \textit{index-based} geometric augmentation is suitable for NDA. First, the geometric transformation only changes the position, and therefore the augmented data is still maintained in event-stream. Second, assume an original event which has $\mathrm{log}V(t) - \mathrm{log}V(t-1) > \alpha$ is still generated in the case of geometric augmentation $f(\mathrm{log}V(t)) - f(\mathrm{log}V(t-1))$. The only difference is position change which can be effectively predicted by the left hand side of Eq.~(\ref{eq_condition}).
For example, rotating a generated set of DVS data and generating a rotated set of DVS data have the same output. 
Therefore, our NDA consists of several geometric augmentations:

\noindent{\bftab Horizontal Flipping.} Flipping is widely used in computer vision tasks. It turns the order of horizontal dimension $w$. Note that for DVS data, the polarity and the time dimension are kept intact. We set the probability of flipping to 0.5. 

\noindent{\bftab Rolling.} Rolling means randomly shifting the geometric position of the DVS image. Similar to a bit shift, rolling can move the pixels left or right in the horizontal dimension. In the vertical dimension, we can also shift the pixels up or down. Note that both circular and non-circular shifts are acceptable since DVS image borders are likely to be 0. Each time we apply rolling, the horizontal shift value $a$ and the vertical shift value $b$ are sampled from an integer uniform distribution $\mathcal{U}(-c,c)$, where $c$ is a hyper-parameter.

\noindent{\bftab Rotation.} The DVS data will be randomly rotated in the clockwise or counterclockwise manner. Similar to rolling, each time we apply rotation, the degree of rotation is sampled from a uniform distribution $\mathcal{U}(-d, d)$, where positive degree means clockwise rotation and vice versa. $d$ is a hyperparameter.

\noindent{\bftab Cutout.} Cutout was originally proposed in~\cite{devries2017cutout} to address overfitting. This method randomly erases a small area of the image, with a similar effect of dropout~\cite{srivastava2014dropout}. Mathematically, first, a square is generated with random size and random position. The side length is sampled from an integer uniform distribution $\mathcal{U}(1, e)$, and then a center inside the image is randomly chosen to place the square. All pixels inside the square are masked off. 

\noindent{\bftab Shear.} Shear mapping originated from plane geometry. It is a linear map that displaces each point in a fixed direction~\cite{Shearmapping}. Mathematically, a horizontal shear (or ShearX) is a function that takes point $(x, y)$ to point $(x+my,y)$. All pixels above $x$-axis are displaced to the right if $m>0$, and to the left if $m<0$. Note that we do not use vertical shear (ShearY) following prior arts~\cite{cubuk2018autoaugment,cubuk2020randaugment}. The shear factor $m$ is also sampled from some uniform distribution $\mathcal{U}(n, n)$.

\noindent{\bftab CutMix.} Proposed in~\cite{yun2019cutmix}, CutMix is an effective linear interpolation of two input data and labels. Consider two input data samples $\bm{x}_{N1}$ and $\bm{x}_{N2}$ and their corresponding label $\bm{y}_1,\bm{y}_2$. CutMix is formulated as
\begin{equation}
    \tilde{\bm{x}}_n = \bm{Mx}_{N1}+\bm{(1-M)x}_{N2},\ \  \tilde{\bm{y}} = \beta\bm{y}_1 + (1-\beta)\bm{y}_2,
\end{equation}
where $\bm{M}$ is a binary mask (similar to the one used in Cutout), and $\beta$ is the ratio between the area of one and the area of zero in that mask.

\begin{figure}[t]
\noindent\begin{minipage}{\textwidth}
\begin{minipage}{0.68\textwidth}
\centering
\captionof{table}{Look-up table for exchanging $N$ with augmentation hyper-parameters.}
\begin{tabular}{l  c  c  c  c}
    \toprule
{\bftab$\boldsymbol{N}$} & {\bftab Rolling} $c$ & {\bftab Rotation} $d$ & \bftab{Cutout} $e$ & {\bftab Shear} $n$ \\
\noalign{\smallskip}
\hline
\noalign{\smallskip}
1 & 3 & 15 & 8 & 0.15 \\
2 & 5 & 30 & 16 & 0.30 \\
3 & 7 & 45 & 24 & 0.45 \\
\bottomrule
\end{tabular}
\label{tab_nda_N}
\end{minipage}
\hfill
\begin{minipage}{0.31\textwidth}
\centering
\includegraphics[width=0.9\linewidth]{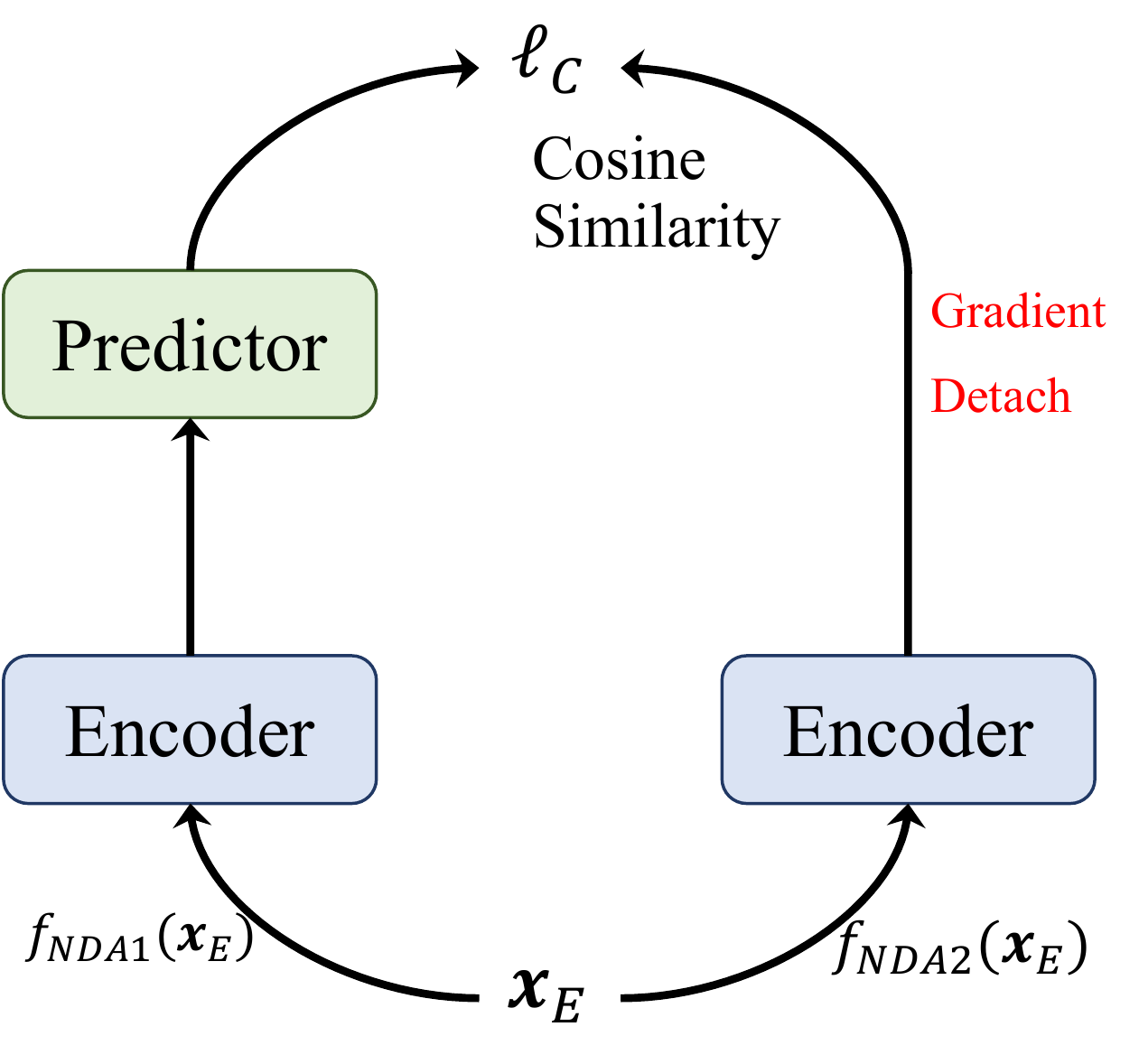}
\captionof{figure}{NDA for SSL-SNN.}
\label{fig_contrast_learning}
\end{minipage}
\end{minipage}
\end{figure}

\subsection{Augmentation Policy}
With each above augmentation, one can generate more training samples. 
In this section, we explore the combination and intensity of the considered augmentations.
We first set flipping and CutMix as the default augmentation, which means they are always enabled. The flipping probability is set to $0.5$ and the CutMix interpolation factor is sampled from $\mathrm{\beta}(1, 1)$. 
Second, for all time step of input data, we randomly sample several augmentations from \{$\mathrm{Rolling}$, $\mathrm{Rotation}$, $\mathrm{Cutout}$, $\mathrm{ShearX}$\}, inspired by prior work~\cite{cubuk2020randaugment,munoz2017deep,zhong2020random}. We define two hyper-parameters: $M$ for the number of augmentations and $N$ for the intensity of augmentations. That is, before each forward pass, we randomly sample $M$ augmentations with $N$-level intensity (the higher the $N$, the greater the difference between augmented and original data). $M$ can be set from 1 to 4. In Table \ref{tab_nda_N}, we describe the corresponding relationship between $N$ and the augmentation hyper-parameters. For instance, using M2N3 policy during a forward pass yields 2 randomly sampled augmentations with $N=3$ intensity as shown in Table \ref{tab_nda_N}. Our algorithm can be simply implemented and can be an add-on extension to existing SNN training methods. 
In the following experiments, we will show how the number of augmentations and the intensity of augmentations impact test accuracy.

\subsection{Application to Unsupervised Contrastive Learning}
Besides improving the supervised learning of SNNs, we show another application of NDA, \textit{i.e.} unsupervised contrastive learning for SNNs.  
Unsupervised contrastive learning~\cite{he2020momentum,chen2020simple,chen2020improved,grill2020bootstrap} is a widely used and well-performing learning algorithm without the requirement of labels.
The idea of contrastive learning is to learn the similarity between paired input images. Each paired output can be either similar or different. 
It would be easy to identify different images, as they are naturally distinct. However, for a  similar image pair, it is required to {augment the same image to different tensors and optimize the network to output the same feature}.  
This task is a simpler task that doesn't require any labels as compared to image classification or segmentation, which makes it perfect for learning some low-level features and transferring the model to some high-level vision tasks. 

In this paper, we implement unsupervised contrastive learning for SNNs based on a recent work Simple Siamese Network~\cite{chen2021exploring} with our proposed NDA, as illustrated in~Fig.~\ref{fig_contrast_learning}.
First, a DVS data input ${\bm{x}_E}$ is augmented into two samples: $f_{\text{NDA1}}({\bm{x}_E})$ and $f_{\text{NDA2}}({\bm{x}_E})$. Our goal is to maximize the similarity of two outputs and one output is further processed by a predictor head to make the final prediction. 
The contrastive loss is defined as the cosine similarity between the predictor output and the encoder output. In contrastive learning, there is a problem called \textit{latent space collapsing}, meaning that the output of a network is the same, irrespective of different inputs which can render the network useless. This collapsing can always yield minimum loss. To address this problem, gradient detach is applied to the branch without predictor. 
It is noteworthy to mention that all the contrastive learning schemes require data augmentation to make model learn invariant features. As a broader impact, this could be helpful when the event camera collects new data that is not easy to be labeled because of raw events. 

After the unsupervised pre-training stage, we drop the predictor and only save the encoder part. This encoder is used for transfer learning, \ie construct a new head (the last fully-connected layer \textit{a.k.a} the classifier) and finetune the model. We will provide transfer results in the experiments section below. 

\section{Experiments}

In this section, we will verify the effectiveness and efficiency of our proposed NDA. In section~\ref{sec_compare}, we compare our model with existing literature. In section~\ref{sec_analysis}, we analyze our method in terms of sharpness and efficiency. In section~\ref{sec_ablation}, we give an ablation study of our methods. In section~\ref{sec_ssl}, we provide the results of unsupervised contrastive learning. 

\noindent{\bftab Implementation details. }We implement our experiments with the Pytorch package. All our experiments are run on 4 GPUs. We use ResNet-19~\cite{he2016deep,zheng2020going} and VGG-11~\cite{simonyan2014very,fang2021incorporating} as baseline models. Note that all ReLU layers are changed to the Leaky integrate-and-fire module and all max-pooling layers are changed to average pooling. We use tdBN~\cite{zheng2020going} as our baseline SNN training method. For our own implementation, we only change the data augmentation part and keep other training hyper-parameters aligned. We use M1N2 NDA configuration for all our experiments. The total batch size is set to 256 and we use Adam optimizer. For all our experiments, we train the model for 200 epochs and the learning rate is set to 0.001 followed by a cosine annealing decay~\cite{loshchilov2016sgdr}. The weight decay is $1e-4$. We verify our method on the following DVS benchmarks:

\noindent{\bftab CIFAR10-DVS~\cite{li2017cifar10}.} CIFAR10-DVS contains 10K DVS images recorded from the original CIFAR10 dataset. We apply a $9:1$ train-valid split (\ie 9k training images and 1k validation images). The resolution is $128\times 128$, we resize all of them to $48\times 48$ in our training and we integrate the event data into 10 frames per sample. 

\noindent{\bftab N-Caltech 101~\cite{orchard2015converting}.} N-Caltech 101 contains 8831 DVS images recorded from the original Caltech 101 dataset. Similarly, we apply $9:1$ train-valid split and resize all images to $48\times 48$. We use the spikingjelly package~\cite{SpikingJelly} to process the data and integrate them into 10 frames per sample. 

\noindent{\bftab N-MNIST~\cite{orchard2015converting}.}
The neuromorphic MNIST dataset is a converted dataset from MNIST. It contains 50K training images and 10K validation images. We pre-process it in the same way as in N-Caltech 101.

\noindent{\bftab N-Cars~\cite{sironi2018hats}.}
Neuromorphic cars dataset is a binary classification dataset with labels either from \textit{cars} or from \textit{background}. It contains 7940 car and 7482 background training samples, 4396 car and 4211 background test samples. We pre-process it in the same way as in N-Caltech 101.

\begin{table}[t]
   \caption{Ablation study: comparison between photometric/color augmentation and geometric augmentation, and comparison with the intensity and the number of the augmentation per data.}
   \centering
   \begin{adjustbox}{max width=\linewidth}
    \begin{tabular}{lcccccc}
       \toprule
    \bftab{Dataset} & {Photo/Color} & {Geo$-$M1N1} & {Geo$-$M1N2} & {Geo$-$M2N2} & {Geo$-$M3N3} \\
    \midrule
    CIFAR10-DVS  & 62.8 & 73.4 & 78.0 & 75.1 & 71.4 \\
    N-Caltech101  &  64.0 & 74.4 & 78.6 & 72.7 & 65.1 \\
   \bottomrule
   \end{tabular}
   \end{adjustbox}
\label{tab_ablation}
\end{table}

\subsection{Ablation Study}
\label{sec_ablation}
{\bftab Augmentation Choice.}
In theory, photometric \& color augmentations are not supposed to be used for DVS data. To verify this, we {\bftab compulsorily cast ColorJitter and GaussianBlur augmentation to the DVS data} (note that the data is not event-stream after these augmentations) and compare the results with geometric augmentation. The results are shown in Table~\ref{tab_ablation} (all entries are results trained with ResNet-19). We find that photometric and color augmentation (Jitter + GaussianBlur) performs much worse than geometric augmentation, regardless of the dataset. This confirms our analysis that \textit{value-based} augmentations are not suitable for NDA. 

\noindent{\bftab Augmentation Policy.} We also test the augmentation intensity as well as the number of the augmentation. In Table~\ref{tab_ablation}, we show that the intensity of the augmentation satisfies some bias-variance trade-off. 
The augmentation can become neither too simple so that the data is not diverse enough, nor too complex so that the data does not contain useful event information.

\subsection{Comparison with Existing Literature}
\label{sec_compare}

\begin{table}[t]
   \caption{Accuracy comparison with different methods on CIFAR10-DVS, N-Caltech 101, N-Cars, we use tdBN in our model. Acc. is referred as the top-1 accuracy. }
   \centering
   \begin{adjustbox}{max width=\linewidth}
   \begin{tabular}{llcccccc}
   \toprule 
   \multirow{2}{4em}{\bftab{Method}} &\multirow{2}{4em}{\bftab{Model}}  & \multicolumn{2}{c}{\bftab{CIFAR10-DVS}} & \multicolumn{2}{c}{\bftab{N Caltech-101}} & \multicolumn{2}{c}{\bftab{N-Cars}} \\
   \cmidrule(l{2pt}r{2pt}){3-4}\cmidrule(l{2pt}r{2pt}){5-6}\cmidrule(l{2pt}r{2pt}){7-8}
    & & {$T$ Step} & {Acc.} & {$T$ Step} & {Acc.} & {$T$ Step} & {Acc.} \\
   \midrule
   HOTS~\cite{lagorce2016hots} & N/A & N/A & 27.1 & N/A & 21.0 & 10 & 54.0 \\ 
   Gabor-SNN~\cite{sironi2018hats} & 2-layer CNN & N/A &  28.4 & N/A & 28.4 & - & - \\ 
   HATS~\cite{sironi2018hats} & N/A & N/A & 52.4 & N/A & 64.2 & 10 & 81.0 \\ 
   DART~\cite{ramesh2019dart} & N/A & N/A & 65.8 & N/A & 66.8& - & - \\
    CarSNN~\cite{viale2021carsnn} & 4-layer CNN & - & -  & - & - & 10 & 77.0 \\
    CarSNN~\cite{viale2021carsnn} & 4-layer CNN$^2$ & - & -  & - & - & 10 & 86.0 \\
   BNTT~\cite{kim2020revisiting} & 6-layer CNN & 20 & 63.2 & - & - & - & - \\
   Rollout~\cite{kugele2020efficient} & VGG-16 & 48 & 66.5 & - & - & - & - \\
   SALT~\cite{kim2021optimizing} &  VGG11 & 20 & 67.1 & 20 & 55.0 & - & -\\ 
   LIAF-Net~\cite{wu2021liaf} & VGG-like & 10 & 70.4 & - & - & - & -\\ 
  tdBN~\cite{zheng2020going} & ResNet-19$^1$ & 10 & 67.8 & - & -& - & - \\
   PLIF~\cite{fang2021incorporating} & VGG-11$^2$ & 20 & 74.8 & - & - & - & -\\
   \midrule
   tdBN (w/o NDA)$^3$ & ResNet-19$^1$ & 10 & 67.9 & 10 & 62.8 & 10 & 82.4 \\
   tdBN (w/. NDA) & ResNet-19$^1$ & 10 & {\bftab 78.0} & 10 & {\bftab 78.6} & 10 & {\bftab 87.2} \\
   \midrule
   tdBN (w/o NDA)$^3$ & VGG-11 & 10 & 76.2 & 10 & 67.2 & 10 &  84.4 \\
   tdBN (w/. NDA) & VGG-11 & 10 & {\bftab 79.6} & 10 & {\bftab 78.2} & 10 & {\bftab 90.1} \\
   tdBN (w/o NDA)$^3$ & VGG-11$^2$ & 10 & 76.3 & 10 & 72.9 & 10 & 87.4 \\
   tdBN (w/. NDA) & VGG-11$^2$ & 10 & {\bftab 81.7} & 10 & {\bftab 83.7} & 10 & {\bftab 91.9} \\
   \bottomrule
   \multicolumn{8}{l}{$^1$ Quadrupled channel number, $^2\ 128\times 128$ resolution, $^3$ Our implementation.}
   \end{tabular}
\label{tab_cifar}
\end{adjustbox}
\end{table}

We first compare our NDA method on the CIFAR10-DVS dataset. The results are shown in Table~\ref{tab_cifar}. We compare our method with Gabor-SNN, Streaming rollout, tdBN, and PLIF~\cite{sironi2018hats,liu2020effective,wu2019direct,kugele2020efficient,zheng2020going,fang2021incorporating}.
Among these baselines, tdBN and PLIF achieve better accuracy. We reproduce tdBN with our own implementations. When training with NDA, we use the best practice M1N2 for sampling augmentations. 
With NDA, our ResNet-19 reaches 78.0\% top-1 accuracy, outperforming the baseline without augmentation by a large margin. After scaling the data to a resolution of 128, our method gets 81.7\% accuracy on VGG-11 model. 
We also compare our method on N-Caltech 101 dataset. There aren't many SNN works on this dataset. Most of them are non-neural network based methods using hand-crafted processing~\cite{lagorce2016hots,sironi2018hats,ramesh2019dart}. NDA can obtain a high accuracy network with 15.8\% absolute accuracy improvement over the baseline without NDA. 
Our VGG-11 even reaches 83.7\% accuracy with full 128$\times$128 resolution. 
This indicates that only improving the network training strategy is less effective.
Next, we test our algorithm on the N-Cars dataset. The primary comparison work is CarSNN~\cite{viale2021carsnn} which optimizes a spiking network and deploys it on the Loihi chip~\cite{davies2018loihi}. tdBN trained ResNet-19 achieve 82.4\% using $48\times48$ resolution. Simply applying NDA improves the accuracy by 4.8\% without any additional modifications. When training with full resolution, we have 4.5\% accuracy improvement with VGG-11.

Finally, we validate our method on the N-MNIST dataset (as shown in Table \ref{tab_nmnist}). N-MNIST is harder than the original MNIST dataset, but most baselines get over 99\% accuracy. We use the same architecture as PLIF. Our NDA uplifts the model by 0.12\% in terms of accuracy. Although this is a marginal improvement, the final test accuracy is quite close to the 100\% mark.

\begin{figure}[t]
\noindent\begin{minipage}{\textwidth}
\begin{minipage}{0.49\textwidth}
\centering
\captionof{table}{N-MNIST accuracy comparison with different methods, we use tdBN in our model.}
   \begin{adjustbox}{max width=\linewidth}
   \begin{tabular}{llcc}
   \toprule 
   {\bftab{Method}} & \bftab{Model} & \bftab{Time Step} & \bftab{Top-1 Acc.} \\
   \midrule
   Lee \etal~\cite{lee2016training} & 2-layer CNN & N/A & 98.61 \\
   SLAYER~\cite{shrestha2018slayer} & 3-layer CNN & N/A & 99.20 \\
   Gabor-SNN~\cite{sironi2018hats} & 2-layer CNN & N/A &  83.70 \\ 
   HATS~\cite{sironi2018hats} & N/A & N/A & 99.10 \\
   STBP~\cite{wu2019direct} & 6-layer CNN & 10 & 99.53 \\
   PLIF~\cite{fang2021incorporating} & 4-layer CNN & 10 & 99.61 \\
   \midrule
   tdBN (w/o NDA) & 4-layer CNN & 10 & 99.58 \\
   tdBN (w/. NDA) & 4-layer CNN & 10 & \bftab{99.70}\\
   \bottomrule
   \end{tabular}
\label{tab_nmnist}
\end{adjustbox}
\end{minipage}
\hfill
\begin{minipage}{0.50\textwidth}
\centering
\captionof{table}{Comparison with EventDrop~\cite{gu2021eventdrop} using ANNs for DVS datasets. EventDrop models are ImageNet pre-trained.}
   \begin{adjustbox}{max width=\linewidth}
   \begin{tabular}{llcc}
   \toprule 
   {\bftab{Model}} & \bftab{Method} & \bftab{N-Caltech 101} & \bftab{N-Cars} \\
   \midrule
   \multirow{4}{5.0em}{ResNet-34} & Baseline~\cite{gu2021eventdrop}$^1$ & 77.4 & 91.8\\
   & EventDrop~\cite{gu2021eventdrop}$^1$ & 78.2 & 94.0\\
   & Ours (w/o NDA)  & 67.7 & 90.5\\
   & Ours (w/. NDA) & \bftab{81.2} & \bftab{95.5}\\
   \midrule
   \multirow{4}{5.0em}{VGG-19} &  Baseline~\cite{gu2021eventdrop}$^1$ & 72.3 & 91.6\\
   & EventDrop~\cite{gu2021eventdrop}$^1$ & 75.0 & 92.7\\
   & Ours (w/o. NDA) & 65.4 &  90.3\\
   & Ours (w/. NDA) & \bftab{82.8} & \bftab{94.5}\\
   \bottomrule
\end{tabular}
\label{tab_ann}
\end{adjustbox}
\end{minipage}
\end{minipage}

\end{figure}

\noindent{\bftab Evaluating NDA on ANN.} We also evaluate if our method is applicable to ANNs. We mainly compare our method with EventDrop on frame data~\cite{gu2021eventdrop}, another work that augments the event data by random dropping some events (like Cutout used in NDA). In Table~\ref{tab_ann}, we find NDA can greatly improve the performance of ANN. For VGG-19, our NDA can boost the accuracy to 82.8\%. For N-Cars dataset, we adopt the same pre-processing as EventDrop, \ie using $8:2$ train-validation split to optimize the model, and our method attains 1.8\% higher accuracy even without any ImageNet pre-training. 

\subsection{Analysis}

\label{sec_analysis}
{\bftab Model Sharpness. }
Apart from the test accuracy, we can measure the \textit{sharpness} of a model to estimate its generalizability. A sharp loss surface implies that a model has a higher probability of mispredicting unseen samples. In this section, we will use two simple and popular metrics: (1) Hessian spectra and (2) Noise injection. 
Hessian matrix is the second-order gradient matrix. 
The second-order gradient contains the curvature information of the model. 
Here, we measure the $\mathrm{topk}$ Hessian eigenvalues and the trace to estimate the sharpness.
Low eigenvalue and low trace lead to low curvature and better performance. 
We compute the 1st, 5th eigenvalues as well as the trace of the Hessian in ResNet-19 using PyHessian~\cite{yao2020pyhessian}. The model is trained on CIFAR10-DVS with and without NDA. We summarize the results in Table~\ref{tab_hessian}. We show that the Hessian spectra of the model trained with NDA is significantly lower than that without NDA. Moreover, the trace of the Hessian also satisfies this outcome. 

\definecolor{myred}{RGB}{178, 34, 35}
\definecolor{myblue}{RGB}{72, 131, 181}

Another way to measure the sharpness is noise injection. We randomly inject Gaussian noise sampled from $\mathcal{N}(0, {\gamma})$ into the model weights, where $\gamma$ is a hyper-parameter controlling the range of the noise. We run 5 times for each $\gamma$ and record the mean \& standard deviation, which are summarized in Fig.~\ref{fig_noise}. It can be seen that the model with NDA is much more robust to the noise we imposed on the weights. For example, when we set the noise to $\mathcal{N}(0, 0.01)$, the model with NDA only suffers a slight 1.5\% accuracy drop, while the model without NDA has a 19.4\% accuracy drop. More evidently, when casting $\mathcal{N}(0, 0.02)$ noise, our NDA model suffers 11.3\% accuracy decline, while the counterpart suffers a drastic 43.4\% accuracy decline. This indicates that NDA serves a similar effect with regularization technique.

\noindent{\bftab Algorithm Efficiency.}
Our NDA is efficient and easy to implement. To demonstrate its efficiency, we hereby report the time cost of NDA. We use Intel XEON(R) E5-2620 v4 CPU to test the data loader. When loading CIFAR10-DVS, the CPU expends additional 15.2873 seconds in applying NDA to 9000 DVS training images. The average cost basically amounts to 1.7ms per image. 

\begin{figure}[t]
\noindent\begin{minipage}{\textwidth}
\begin{minipage}{0.49\textwidth}
\centering
\includegraphics[width=\linewidth]{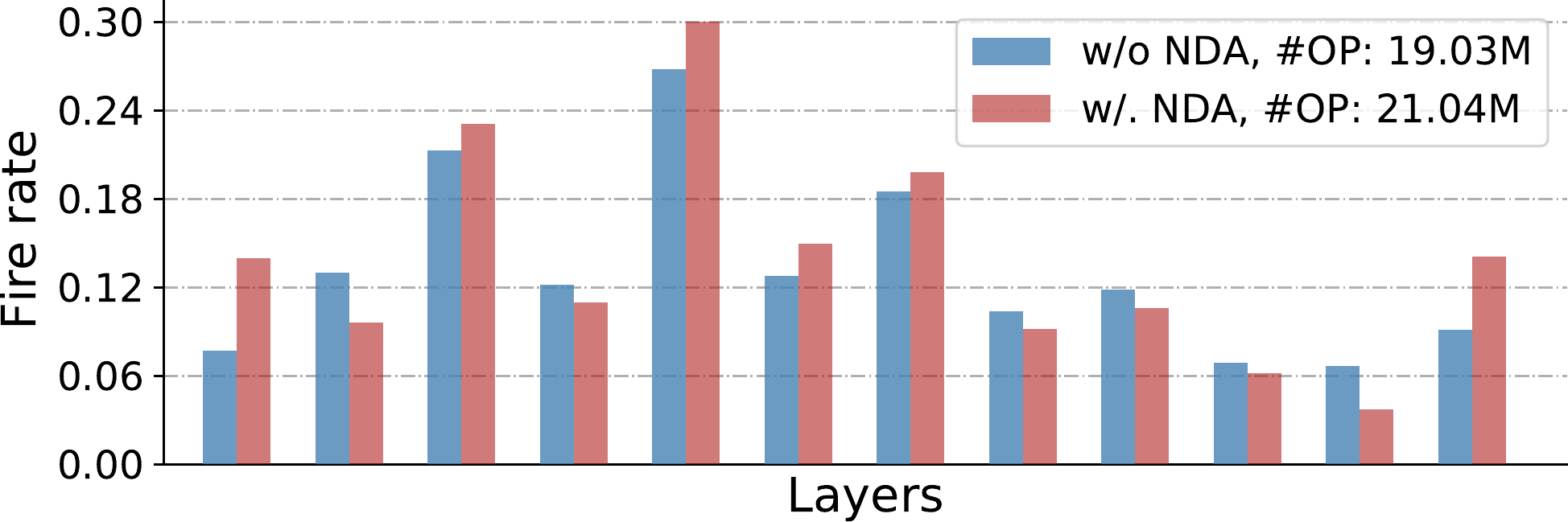}
\captionof{figure}{Fire Rate of several LIF layers of ResNet-19 trained on CIFAR10-DVS.}
\label{fig_fire}
\end{minipage}
\hfill
\begin{minipage}{0.49\textwidth}
\centering
\captionof{table}{Ablation Study: the effect of validation dataset. }
\centering
\begin{adjustbox}{max width=\linewidth}
\begin{tabular}[b]{l  c  c} 
\toprule
{{\bftab Method}} & {\bftab Acc. No Valid.} & {\bftab Acc. 15\% Valid.} \\
\noalign{\smallskip}
\hline
\noalign{\smallskip}
PLIF \cite{fang2021incorporating} & 74.8 & 69.0 \\ 
\noalign{\smallskip}
\hline
\noalign{\smallskip}
{\textcolor{myblue}{\bftab w/o NDA}} & 63.4 & 58.5 \\
{\textcolor{myred}{\bftab w/. NDA}} & 78.0 & {\bftab 74.4}\\
\bottomrule
\end{tabular}
\end{adjustbox}
\label{tab_val}
\end{minipage}
\end{minipage}
\end{figure}

We also estimate the energy consumption of the model using NDA. Specifically, we use the number of operations and roughly assume energy consumption is linearly proportional to the number of operations~\cite{rathi2020diet,zheng2020going}. We do not count the operation if the spike does not fire. Thus, the number of operations can be computed as $FireRate \times MAC$. In Fig.~\ref{fig_fire}, we visualize the $Fire Rate$ of several layers in the trained model with and without NDA. We can find that the $Fire Rate$ is similar (note, NDA is slightly higher) and the final number of operations are very close in both cases. Model with NDA only expends a 10\% higher number of operations than the model without NDA, demonstrating the efficiency of NDA.

\noindent{\bftab Training with validation dataset. }Following~\cite{fang2021incorporating}, we test our algorithm on a challenging task. We take 15\% of the data from the training dataset to build a \emph{validation dataset}. Then, we report the \emph{test accuracy} when the model reaches best \emph{validation accuracy}. We run experiments on CIFAR10-DVS with ResNet-19. In this case (Table \ref{tab_val}), our NDA model yields 16.3\% accuracy improvement over a model without NDA. 
\begin{figure}[t]
\noindent\begin{minipage}{\textwidth}
\begin{minipage}{0.32\textwidth}
\centering
\includegraphics[width=0.9\linewidth]{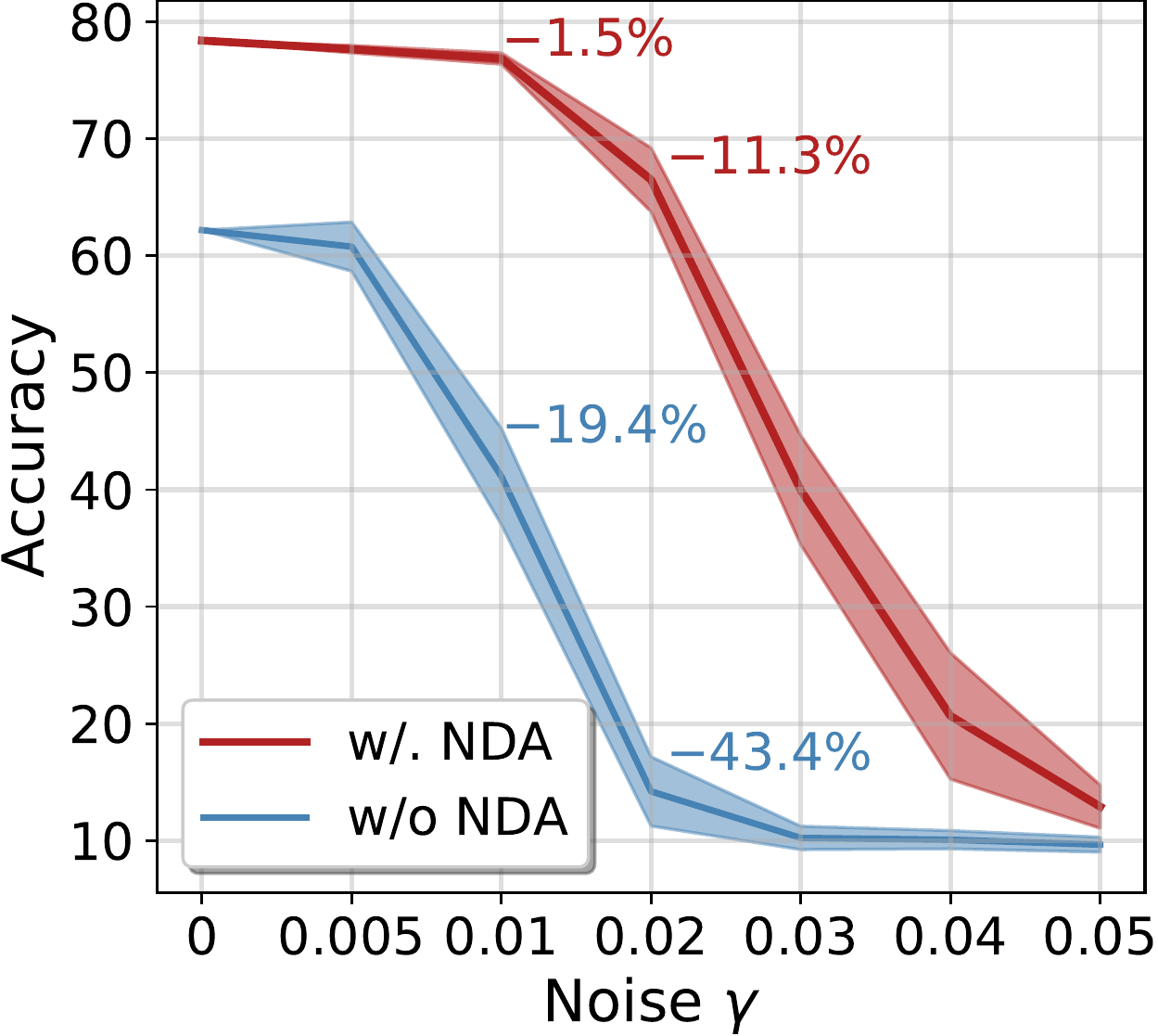}
\captionof{figure}{Comparison of robustness under Gaussian noise injected to weights.}
\label{fig_noise}
\end{minipage}
\hfill
\begin{minipage}{0.63\textwidth}
\centering
\captionof{table}{Hessian spectra comparison. $\lambda_1, \lambda_5, Tr$ refer to the 1st, 5th highest eigenvalue and the trace of the Hessian matrix. We record the model at 100, 200, 300 epochs. (The lower the Hessian spectrum, the flatter the converged minimum is). }
\label{tab_hessian}
\begin{adjustbox}{max width=\linewidth}
\begin{tabular}{lcccccc}
\toprule  
\multirow{2}{3em}{\bftab{Epoch}}& \multicolumn{3}{c}{\textcolor{myblue}{\bftab{w/o NDA}}} & \multicolumn{3}{c}{\textcolor{myred}{\bftab{w/. NDA}}} \\
\cmidrule(l{2pt}r{2pt}){2-4} \cmidrule(l{2pt}r{2pt}){5-7}
 & $\lambda_1$ & $\lambda_5$ & $Tr$ & $\lambda_1$ & $\lambda_5$ & $Tr$\\
\midrule 
100 & 910.7 & 433.4 & 6277 & 424.3 & 73.87 & 1335 \\
200 & 3375 & 1416 & 21342  & 516.4 & 155.3 & 1868 \\
300 & 3404 & 1686 & 20501  & 639.7 & 187.5 & 2323 \\
\bottomrule
\end{tabular}
\end{adjustbox}
\end{minipage}
\end{minipage}
\end{figure}

\subsection{Unsupervised Contrastive Learning}
\label{sec_ssl}
In this section, we test our NDA algorithm with unsupervised contrastive learning. Usually, the augmentation in unsupervised contrastive learning requires stronger augmentation than that in supervised learning~\cite{chen2020simple}. Thus we use NDA-M3N2 for augmentation. We pre-train a VGG-11 on the N-Caltech 101 dataset with SimSiam learning (cf. Section 4.3, Fig. \ref{fig_contrast_learning}). N-Caltech 101 contains more visual categories and can be a good pre-training dataset.
In this experiment, we use the original $128\times 128$ resolution DVS data, and use the simple-siamese method to train the network for 600 epochs. The batch size is set to 256. 

After pre-training, we replace the predictor with a zero-initialized fully-connected classifier and finetune the model for 100/300 epochs on CIFAR10-DVS. We also add supervised pre-training with NDA and no pre-training (\ie directly train a model on downstream task) as our baselines. We show the transferability results in Table~\ref{tab_ssl}. Interestingly, we find supervised pre-training has no effect on transfer learning with such DVS datasets, which is different from conventional natural image classification transferability results. This may be attributed to the distance between dataset domains that is larger in DVS than that in RGB dataset. The model pre-trained on N-Caltech 101 with supervised learning only achieves 80.9\% accuracy, which is even 0.8\% lower than the no pre-training method. The unsupervised learning with NDA achieves significantly better transfer results: finetuning only for 100 epochs takes the model to 80.8\% accuracy on CIFAR10-DVS, and 300 epochs finetuning yields 82.1\% test accuracy, establishing a new state of the art.

\begin{table}[t]
\caption{Unsupervised transfer learning results on CIFAR10-DVS. All the model is pre-trained on N-Caltech 101. }
\centering
\begin{adjustbox}{max width=\linewidth}
\begin{tabular}[b]{l l c} 
\toprule 
{\bftab Pre-training Method} \:\:\:& {\bftab Finetuning Method} \:\:\: & {\bftab Test Acc.} \\
\midrule
No Pre-training & Train@300 & 81.7 \\
Supervised & Finetune@100 &  77.4 \\
Supervised & Finetune@300 &  80.9 \\
Unsupervised & Finetune@100 & 80.8 \\
Unsupervised & Finetune@300 & {\bftab 82.1} \\
\bottomrule
\end{tabular}
\end{adjustbox}
\label{tab_ssl}
\end{table}

\section{Conclusions}

We introduce the Neuromorphic Data Augmentation technique (NDA), a simple yet effective method that improves the generalization of SNNs on event-based data. 
NDA allows users to generate new high-quality event-based data instances that force the model to learn invariant features.     
Furthermore, we show that NDA acts like regularization that achieves an improved bias-variance trade-off.
Extensive experimental results validate that 
NDA is able to find a flatter minimum with the higher test accuracy and enable unsupervised pre-training for transfer learning. However, the current NDA lacks value-based augmentations for events, which may be realized by logical operations and studied in the future. 

\nocite{christensen20222022}

\noindent\textbf{Acknowledgment.}
This work was supported in part by C-BRIC, a JUMP center sponsored by DARPA and SRC, Google Research Scholar Award, the National Science Foundation (Grant\#1947826), TII (Abu Dhabi), and the DARPA AI Exploration (AIE) program. 

%
%
\bibliographystyle{splncs04}
\bibliography{egbib}
\end{document}